\documentclass[lettersize,journal]{IEEEtran}
\usepackage{amsmath,amsfonts}
\usepackage{algorithm}
\usepackage{algpseudocode}
\usepackage{array}
\usepackage[]{subfig}
\usepackage{textcomp}
\usepackage{stfloats}
\usepackage{url}
\usepackage{verbatim}
\usepackage{graphicx}
\usepackage{cite}
\usepackage{hyperref}
\usepackage{bm}
\usepackage{makecell}
\usepackage{footnote}
\usepackage{float}
\usepackage{subfig}
\usepackage{bbding}
\usepackage{ragged2e}
\usepackage{amsthm}
\usepackage{dsfont}
\usepackage{lipsum} 
\usepackage{booktabs}
\usepackage{multirow}
\usepackage{xcolor}
\usepackage{makecell}
\usepackage[normalem]{ulem}
\useunder{\uline}{\ul}{}

\usepackage{enumitem} 
\usepackage{pifont}

\makeatletter

\newcommand{\Rmnum}[1]{\expandafter\@slowromancap\romannumeral #1@}
\makeatother

\theoremstyle{definition}

\usepackage{cuted}

\usepackage[switch]{lineno}

\begin{document}

\title{BearingPGA-Net: A Lightweight and Deployable Bearing Fault Diagnosis Network via Decoupled Knowledge Distillation and FPGA Acceleration}

\author{Jing-Xiao~Liao$^{1,2}$, \textit{Student Member, IEEE}, Sheng-Lai Wei$^{1,3}$, Chen-Long Xie$^{1}$, Tieyong Zeng$^{4}$, Jinwei Sun$^{1}$, Shiping Zhang$^{1*}$, Xiaoge Zhang$^{2*}$, Feng-Lei Fan$^{4*}$

\thanks{*Shiping Zhang (spzhang@hit.edu.cn), Xiaoge Zhang (xiaoge.zhang@polyu.edu.hk), and Feng-Lei Fan (hitfanfenglei@gmail.com) are co-corresponding authors.}
\thanks{$^{1}$School of Instrumentation Science and Engineering, Harbin Institute of Technology, Harbin, China.}
\thanks{$^{2}$Department of Industrial and Systems Engineering, The Hong Kong Polytechnic University, Hong Kong SAR.}
\thanks{$^{3}$Analog Devices, Inc., Shanghai, China.}
\thanks{$^{4}$Department of Mathematics, The Chinese University of Hong Kong, Hong Kong SAR.}}


\maketitle

\begin{abstract}

Deep learning has achieved remarkable success in the field of bearing fault diagnosis. However, this success comes with larger models and more complex computations, which cannot be transferred into industrial fields requiring models to be of high speed, strong portability, and low power consumption. In this paper, we propose a lightweight and deployable model for bearing fault diagnosis, referred to as BearingPGA-Net, to address these challenges. Firstly, aided by a well-trained large model, we train BearingPGA-Net via decoupled knowledge distillation. Despite its small size, our model demonstrates excellent fault diagnosis performance compared to other lightweight state-of-the-art methods. Secondly, we design an FPGA acceleration scheme for BearingPGA-Net using Verilog. This scheme involves the customized quantization and designing programmable logic gates for each layer of BearingPGA-Net on the FPGA, with an emphasis on parallel computing and module reuse to enhance the computational speed. To the best of our knowledge, this is the first instance of deploying a CNN-based bearing fault diagnosis model on an FPGA. Experimental results reveal that our deployment scheme achieves over 200 times faster diagnosis speed compared to CPU, while achieving a lower-than-0.4\% performance drop in terms of F1, Recall, and Precision score on our independently-collected bearing dataset. Our code is available at \url{https://github.com/asdvfghg/BearingPGA-Net}.

\end{abstract}


\section{Introduction}
\IEEEPARstart{R}{otating} machines, including turbines, pumps, compressors, fans, etc., are widely used in industrial fields~\cite{qiao2015survey, 10042467, boudiaf2016comparative}. Rolling bearings, whose primary roles are to support the whole machine and reduce friction, are crucial components in rotating machines. Statistics from authoritative institutes reveals that bearing failure accounts for approximately 45\%–70\% of all mechanical faults~\cite{bonnett2008increased, benbouzid2000review}. Therefore, detecting bearing failure is an essential task in industry and civil fields. A timely and accurate detection can greatly enhance the reliability and efficiency of bearings, thereby reducing the enormous economic loss.

Since bearing failures often exhibit unique abnormal vibration~\cite{smith2015rolling, rauber2014heterogeneous, qiao2015survey2}, the most common approach to diagnosing bearing faults is to install an acceleration sensor on mechanical surfaces to measure vibration signals, and then a diagnosis algorithm is applied to detect faulty signals.
Over the past decade, deep learning methods, represented by convolutional neural networks (CNNs) and attention-based models, have been dominating the problem of bearing fault diagnosis~\cite{liao2023attention, 9666871, wang2022attention, 9851479}. Despite the significant success achieved by deep learning methods, the growing model size renders them impractical in industrial settings, since deploying deep learning models often demands high-performance computers. But a factory usually has multiple rotating machines. For instance, in nuclear power plants, there are numerous rotating machines such as seawater booster pumps and vacuum pumps that require monitoring of their vibration signals~\cite{10158933}. Thus, installing computers for each machine will not only undesirably occupy space but also impede production due to excessive circuit connections. 

In fact, the field programmable gate array (FPGA), a class of low-power consumption, high-speed, programmable logic devices, is widely used in industrial fields. FPGA has high-speed signal processing capabilities such as time-frequency analysis, which can enhance the accuracy and real-time performance of fault detection~\cite{jamshidpour2014fpga,dick2000fpga,choi2003energy}. By leveraging its parallel computing capabilities, FPGA dramatically increases the speed and energy efficiency, achieving over a thousand times improvement than a digital signal processor (DSP) in fault detection execution~\cite{liu2019real}. However, the FPGA suffers from limited storage space and computational resources, which brings huge trouble in bearing fault diagnosis, as it needs to process long signal sequences in a timely manner. 
In brief, to deploy fault diagnosis in practical scenarios, we need to address two challenges: designing lightweight but well-performed deep learning models and deploying these models on embedded hardware such as FPGAs without compromising the performance too much.

Along this line, researchers have put forward lightweight deep models for bearing fault diagnosis, aiming to strike a balance between diagnostic accuracy and model complexity~\cite{yao2020lightweight,fang2021clformer,10172025,zhong2022bearing}. Nevertheless, these lightweight models have not been really deployed on the embedded hardware, and whether or not the good performance can be delivered as expected remains uncertain. On the other hand, two types of hardware have been utilized to deploy CNNs, System-on-Chip (SoC) FPGA and Virtex series FPGA. The former, exemplified by ZYNQ series chips which integrate an ARM processor with an FPGA chip, have been employed to deploy a limited number of bearing fault diagnosis methods~\cite{toumi2022fpga,ji2022neural}. These approaches rely on the PYNQ framework, which translates the Python code into FPGA-executed Verilog through the ZYNQ board. While the PYNQ framework reduces development time, the FPGA resources available for CNN acceleration are limited, reducing the processing time compared to FPGA boards. As for the latter, some prior researchers have designed FPGA-based accelerators for CNNs and successfully deployed CNNs on Virtex series FPGA~\cite{qiu2016going,mittal2020survey,zhang2015optimizing}. However, these FPGAs, capable of deploying multi-layer CNNs, cost ten times higher than commonly used mid-range FPGAs (Kintex series). Consequently, widespread deployment is hindered due to excessive costs. To resolve the tension between model compactness and deployment, here we propose BearingPGA-Net: a lightweight and deployable bearing fault diagnosis network via decoupled knowledge distillation and FPGA acceleration.

First, we apply the decoupled knowledge distillation (DKD) to build a lightweight yet well-performed network, named as BearingPGA-Net. Knowledge distillation (KD) is a form of model compression that transfers knowledge from a large network (teacher) to a smaller one (student)~\cite{hinton2015distilling,gou2021knowledge}, which is deemed as more effective than directly prototyping a small network. Knowledge distillation can be categorized into response-based KD (logit KD) and feature-based KD. Albeit previous studies have demonstrated the superiority of feature-based KD over logit KD for various tasks~\cite{romero2014fitnets, wang2021knowledge}, we select the logit KD. Our student network comprises only a single convolutional layer for deployment, and there are no hidden features for feature-based KD to distill. Recently, a novel approach called decoupled knowledge distillation has been proposed, which reformulates the traditional logit distillation by decomposing the knowledge distillation loss function into two separate functions. The training of these functions can be flexibly balanced based on the task~\cite{zhao2022decoupled}. The decoupling is a strong booster for the performance of logit KD, and we favorably translate it into our task.

Second, we utilize Verilog to implement neural network accelerators. By devising parallel computing and module
reuse, we deploy BearingPGA-Net and enhance its computing speed in a Kintex-7 FPGA. Specifically, we construct basic arithmetic modules with basic multiplication and addition units for the convolutional and fully-connected layers. To cater to the requirements of BearingPGA-Net, we fuse a ReLU and Max-pooling layer to further reduce computation. These computational modules are also translatable to other CNN-based tasks. Moreover, we design a tailored layer-by-layer fixed-point quantization scheme for neural networks, ensuring minimal loss of parameter accuracy in FPGA computations while also cutting the number of parameters by half. Notably, our FPGA framework fully leverages the computational resources of the Kintex-7 FPGA, which is a widely used mid-range FPGA in industrial fields. Compared to previous implementations via SoC FPGA and Virtex FPGA, BearingPGA-Net+Kintex FPGA achieves preferable power consumption, model compactness, and high performance, which is highly scalable in real-world fault diagnosis scenarios. In summary, our contributions are threefold:

{\small $\bullet$} We build BearingPGA-Net, a lightweight neural network tailored for bearing fault diagnosis. This network is characterized by a single convolutional layer and is trained via decoupled knowledge distillation.

{\small $\bullet$} We employ fixed-point quantization to compress the parameters of BearingPGA-Net by 50\% and propose a CNN accelerators scheme, where we utilize parallel computing and module reuse techniques to fully leverage the computational resources of the Kintex-7 FPGA.

{\small $\bullet$} Compared to lightweight competitors, our proposed method demonstrates exceptional performance in noisy environments, achieving an average F1 score of over 98\% on CWRU datasets. Moreover, it offers a smaller model size, with only 2.83K parameters. Notably, our FPGA deployment solution is also translatable to other FPGA boards.

\section{Related Works}

\textbf{1) Lightweight CNNs for bearing fault diagnosis.} Some lightweight networks were proposed for deployment in resource-constrained environments. Yao \textit{et al.} introduced the stacked inverted residual convolution neural network (SIRCNN), comprising one convolutional layer and three inverse residual layers~\cite{yao2020lightweight}. Similarly, Fang \textit{et al.} developed the lightweight efficient feature extraction network (LFEE-NET), which is a feature extraction module conjugated with a lightweight classification module. However, despite the simple structure of the classification network, its feature extraction network is complex~\cite{fang2021clformer}. Other lightweight models incorporated a multi-scale model~\cite{10172025} or a self-attention mechanism~\cite{zhong2022bearing}, demonstrating the superior performance in handling few-shot or cross-domain issues. But these networks have not yet been deployed on embedded devices. Therefore, it remains unclear how much performance will be sacrificed when deployed.

\textbf{2) Bearing fault diagnosis models for FPGAs.} There were only a small number of models successfully deployed on FPGAs. An FPGA-based multicore system was proposed for real-time bearing fault diagnosis using acoustic emission (AE) signals \cite{kang2014fpga}. It designed a high-performance multicore architecture including 64 processing units running on a Xilinx Virtex-7 FPGA to support online processing, and using time-frequency analysis and support vector machine (SVM) for diagnosis~\cite{kang2014fpga}. Toumi implemented an envelope spectrum and multi-layer perceptron (MLP) structure on a ZYNQ-7000 FPGA, achieving around 90\% accuracy in CWRU datasets~\cite{toumi2022fpga}. Ji \textit{et al.} used the knowledge distillation technique to train a single-layer convolutional neural student network and deployed it into a ZYNQ FPGA through parameter quantization, resulting in an approximate 8\% improvement compared to training the network directly ~\cite{ji2022neural}. 

Despite these achievements, the task of deploying fault diagnosis models in FPGAs still have a large room for improvement, as their performance has not yet reached the level of the large model. Our idea is to combine CNNs and signal processing techniques so that it can achieve real-time and high-performance fault diagnosis in industrial fields.

\section{Method}
\begin{figure}[htbp]
\vspace{-0.2cm}
    \centering
    \includegraphics[width=0.9\linewidth]{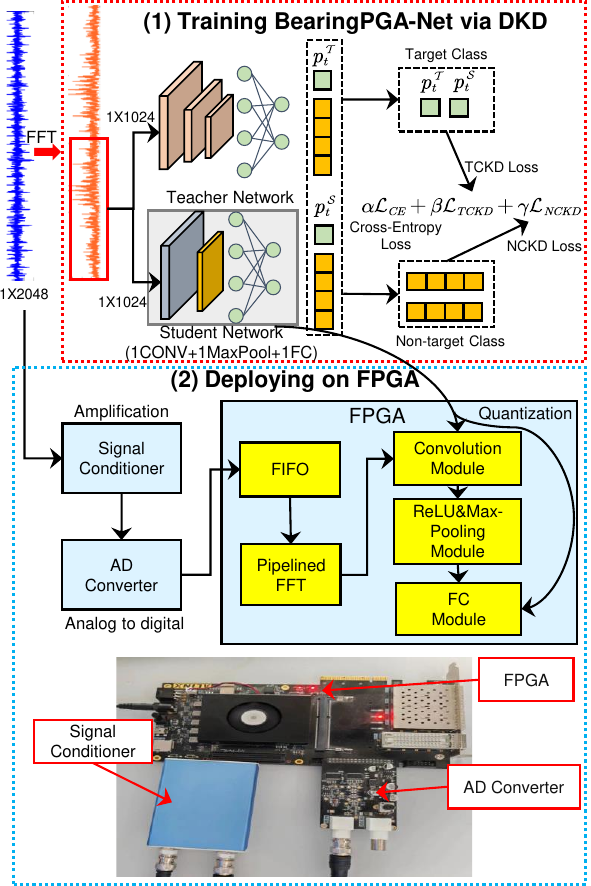}
    \caption{The overall framework for prototyping and deploying BearingPGA-Net. }
    \label{fig:framework}
    \vspace{-0.8cm}
\end{figure}

As depicted in Fig.~\ref{fig:framework}, prototyping BearingPGA-Net follows a two-step pipeline: i) training a lightweight BearingPGA-Net via decoupled knowledge distillation; ii) deploying the BearingPGA-Net into an FPGA. Notably, we devise a layer-by-layer fixed-point quantization method to convert the parameters (weights and bias) of PyTorch's CNN, which are originally in 32-bit floating format, into a 16-bit fixed-point format. Additionally, for online diagnosis, the measured signal is amplified by a signal conditioner and then converted into a digital signal by an Analog-Digital (AD) converter. Subsequently, the FPGA's FIFO (first in first out) performs clock synchronization and buffering, followed by executing convolutional neural network operations. Finally, the diagnosing result is displayed using four LEDs.

\subsection{Training BearingPGA-Net}

The BearingPGA-Net is trained via decoupled knowledge distillation (DKD), which is an approach to training a small model (student) with a well-trained large model (teacher). It forces the student to emulate the output of the teacher. During the training process, the teacher model is trained first. Subsequently, its parameters are freezed to generate the outputs as new labels to train the student model.

\textbf{1) Constructing teacher and student networks.} For the teacher network, we adopt a widely-used 1D-CNN architecture known as the WDCNN~\cite{zhang2017new}, which consists of six CNN blocks and one fully-connected layer. In our implementation, we adjust the number of weights in the fully-connected layer to fit the input data size.

Additionally, we design a one-layer student network specifically for FPGA deployment (BearingPGA-Net). This student network comprises a single 1D-CNN layer, a ReLU activation layer and a Max-Pooling layer, followed by a fully-connected layer mapping the latent features to the logits. The structure information of BearingPGA-Net is shown in Tab. \ref{tab:studentparam}.

\begin{table}[h]
\caption{The structure parameters of BearingPGA-Net.}
\centering
\begin{tabular}{|l|c|c|c|c|c|}
\hline
Layer    & Kernel size & Channel & Output size & Padding & Stride \\ \hline
Conv1d    & 64×4        & 4       & 128×4       & 28      & 8      \\ \hline
ReLU     & -           & -       & 128×4       & -       & -      \\ \hline
MaxPool & 2×1         & 4       & 64×4        & 0       & 2      \\ \hline
Linear  & (256,10)    & -      & 10    & -       & -      \\ \hline
\end{tabular}
\label{tab:studentparam}
\end{table}

\textbf{2) Decoupled knowledge distillation.} 
Despite numerous forms of knowledge distillation, our method adopts the response-based KD, which utilizes the teacher model's logits for knowledge transfer. This is because the limited hardware resource in a low-cost FPGA cannot hold two or more convolutional layers. Then, there are no hidden features for feature-based KD to distill. 

In classical knowledge distillation, soft labels, which are the logits produced by the teacher, are deemed as distilled knowledge \cite{hinton2015distilling}. Soft labels are obtained by the softmax function converting the logits $z_i$ of a neural network into the probability $p_i$ of the $i$-th class:
\begin{equation}
    p_i=\frac{\exp{(z_i/T)}}{\sum_{j=1}^{C}\exp(z_j/T)},
\label{eq:softmax}
\end{equation}
where $C$ represents the number of classes, and $T \in \mathbb{R}^+$ serves as the temperature factor. When the temperature $T$ is higher, the probability distribution over classes becomes smoother. A lower value of $T$ (where $T<1$) sharpens the output, increasing the disparity in probability values of different classes.


For classification, the cross-entropy (CE) loss is adopted to measure the difference between the probability of the predicted label ${p}$ and the ground truth ${y}$:
\begin{equation}
    \mathcal{L}_{CE}(y,p(z,T))=-{\textstyle \sum}_{i=1}^{C}y_i\log p_i,
\end{equation}
and KL-Divergence measures the similarity between the probability labels of the teacher $p^\mathcal{T}$ and the student $p^\mathcal{S}$, 
\begin{equation}
    \mathcal{L}_{KL}(p^\mathcal{T}(z, T),p^\mathcal{S}(z, T)) = -{\textstyle \sum}_{i=1}^{C}p^\mathcal{T}_i\log \frac{p^\mathcal{S}_i}{p^\mathcal{T}_i}.
\end{equation}
KD combines two loss functions:
\begin{equation}
\mathcal{L}_{KD}
=(1-\alpha) \mathcal{L}_{CE}(y,p^\mathcal{S})+\alpha T^2 \mathcal{L}_{KL}(p^\mathcal{T}(z, T),p^\mathcal{S}(z, T)),
\label{eq:kd}
\end{equation}
where $\alpha$ is the scale factor to reconcile the weights of two loss functions, and $T^2$ keeps two loss functions at the same level of magnitude. Combining $\mathcal{L}_{KL}$ and $\mathcal{L}_{KD}$ helps the student get the guidance of the teacher and the feedback from the ground truth. This ensures that the student model learns effectively and reduces the risk of being misled.


However, $\mathcal{L}_{KL}$ implies that all logits from the teacher are transferred to the student on an equal booting. Intuitively, student models should have the ability to filter the knowledge they receive. In other words, knowledge relevant to the current target category should be reinforced, while knowledge far from the target category should be attenuated. The coupling in classical KD harms the effectiveness and flexibility across various tasks. To address this, decoupled knowledge distillation, which factorizes $\mathcal{L}_{KL}$ into a weighted sum of two terms (the target class and non-target class), was proposed~\cite{zhao2022decoupled}. 

Let $p_t$ and $p_{/t}$ be probabilities of the target and non-target classes, respectively. Then, we have 
\begin{equation}
    p_t= \frac{\exp \left( z_t /T \right)}{\sum_{j=1}^C{\exp \left( z_j /T\right)}}, \ p_{/t}= \frac{\sum_{i=1,i\ne t}^{C} \exp \left( z_i /T \right)}{\sum_{j=1}^C{\exp \left( z_j /T\right)}},
\end{equation}
and
\begin{equation}
    \hat{p}_t=\frac{p_t}{p_{/t}}=\frac{\exp \left( z_t /T \right)}{\sum_{i=1,i\ne t}^C{\exp \left( z_i /T\right)}}.
\end{equation}  
Then, $\mathcal{L}_{KL}$ can be factorized as
\begin{equation}
\begin{aligned}
&\mathcal{L}_{KL}(p^\mathcal{T}(z, T),p^S(z, T)) \\
=& -p^\mathcal{T}_t\log \frac{p^\mathcal{S}_t}{p^\mathcal{T}_t} -\sum_{i=1, i \ne t}^{C}p^\mathcal{T}_i\log \frac{p^\mathcal{S}_i}{p^\mathcal{T}_i}\\
=& \underbrace{-p^\mathcal{T}_t\log \frac{p^\mathcal{S}_t}{p^\mathcal{T}_t}-p^\mathcal{T}_{/t}\log \frac{p^\mathcal{S}_{/t}}{p^\mathcal{T}_{/t}}}_{\mathcal{L}_{KL}^{TCKD}}-p_{/t}^\mathcal{T}\underbrace{\sum_{i=1, i \ne t}^{C}\hat{p}^\mathcal{T}_i\log \frac{\hat{p}^\mathcal{S}_i}{\hat{p}^\mathcal{T}_i}}_{\mathcal{L}_{KL}^{NCKD}},\\
\end{aligned}
\end{equation}
where $\mathcal{L}_{KL}^{TCKD}$ denotes the KL loss between
teacher and student's probabilities of the target class, named target class knowledge distillation (TCKD), while $\mathcal{L}_{KL}^{NCKD}$ denotes the KL loss between teacher and student's  probabilities of the non-target classes, named non-target class knowledge distillation (NCKD). The derivation can also be found in \cite{zhao2022decoupled}.

Thus, the entire DKD loss function is
\begin{equation}
\mathcal{L}_{DKD}= (1-\alpha) \mathcal{L}_{CE}(y,p^\mathcal{S})+\alpha T^2 (\beta \mathcal{L}_{KL}^{TCKD} + \gamma  \mathcal{L}_{KL}^{NCKD}),
\label{eq:dkd}
\end{equation}
where $p_{t}$ and $p_{/t}$ are merged into two hyperparameters $\beta$ and $\gamma$ to coordinate the contributions of two terms. In bearing fault diagnosis, encouraging TCKD while suppressing NCKD ($\beta > \gamma$) often leads to performance improvement. By optimizing this decoupled loss, the knowledge acquired by the teacher model is more easily imparted into the student model, thereby improving the student network's performance. 

\subsection{Deploying BearingPGA-Net into FPGA}


The PYNQ framework can directly convert the Python code into FPGA bitstream files, but it is specifically designed for the architecture of a System-on-Chip (SoC) FPGA. This type of FPGA integrates an ARM processor and an FPGA chip into a single package. However, the FPGA in such an SoC has only 50k look-up table (LUT), while other FPGAs have over 200k LUT. As a result, the computation speed of the SOC FPGA is much slower \cite{toumi2022fpga, ji2022neural} than other FPGAs.

In contrast, without bells and whistles, we design a simple yet pragmatic CNN acceleration scheme that directly translates the network computation modules in logic circuitry, which is directly deploying the module into FPGA chips, without the need for ARM chip translation, thereby fully leveraging the FPGA's high-speed computing capabilities. Specifically, we convert the operations of BearingPGA-Net into Verilog and optimize the place and route (P\&R) of the logical connection netlist such as logic gates, RAM, and flip-flops to implement hardware acceleration. The entire process includes register transfer level (RTL) design, simulation and verification, synthesis, implementation (P\&R), editing timing constraints, generation of bitstream files, and the FPGA configuration. Notably, our scheme is also scalable to other FPGA chips such as Virtex-7series.

\textbf{1) Fixed-point quantization.} The original network in Python employs the 32-bit floating-point numbers for high-accuracy calculations. However, these 32-bit numbers consume a significant amount of memory, which cannot allow the deployment of single-layer networks. To address this, we employ fixed-point quantization, which converts the numbers to 16-bit fixed-point format. In this format, the bit-width keeps intact for integers, decimals, and symbols in a floating-point number. The reduction in precision is up to a half of the original, despite a minor decrease in performance. The fixed-point number is expressed as
\begin{equation}
    Q_{X,Y} = \pm {\textstyle \sum}_{i=0}^{X-1}b_i\cdot2^i+{\textstyle \sum}_{j=1}^{Y}b_{X+j}\cdot2^{-j}.
\end{equation}
where $Q_{X,Y}$ is a fixed-point number that contains $X$-bit integers and $Y$-bit decimals, $b_i$ and $b_{X+j}$ are the value on the binary bit. The storage format of fixed-point numbers in an FPGA is $(\pm, b_i, \cdot, b_{X+j})$.


\begin{figure}[h]
    \vspace{-0.4cm}
    \centering
    \includegraphics[width=0.8\linewidth]{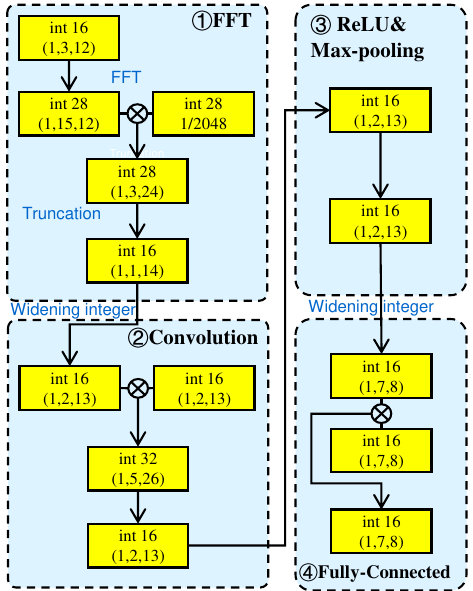}
    \caption{An example of bit-width of integers and decimals in each layer of BearingPGA-Net on FPGA computation, where $(S,X,Y)$ denotes the number of symbol bit, integer bits, and decimal bits, respectively.}
    \label{fig:quantization}
    \vspace{-0.3cm}
\end{figure}

Furthermore, when considering a fixed bit-width, it is crucial to determine the trade-off between the bit-widths of integers and decimals. It is important to have a sufficiently large range to minimize the probability of overflow while ensuring a small enough resolution to reduce quantization error \cite{lin2016fixed}. In this study, a dynamic fixed-point quantization is assigned to each layer of the BearingPGA-Net to preserve the overall accuracy~\cite{williamson1991dynamically, abdelouahab2018accelerating}. The specific bit-width for integers and decimals is determined based on the maximum and minimum values of parameters in each layer, so different models may have various bit-width. The allocation of bit-width of integers and decimals in each layer is illustrated in Fig.~\ref{fig:quantization}.

\textbf{2) The overall workflow.}
As shown in Fig. \ref{fig:implestru}. Parameters are stored in the ROM after quantization, and we implement the BearingPGA-Net accelerator (FFT, convolutional layer, max-pooling conjugated with ReLU, fully-connected layer) using Verilog. The FFT operation is performed by the IP core, and we carefully design logic circuit modules for all operations of the BearingPGA-Net to carry out. Moreover, some specific modules used in FPGAs are described below:

\begin{figure}[htbp]
\vspace{-0.3cm}
    \centering
    \includegraphics{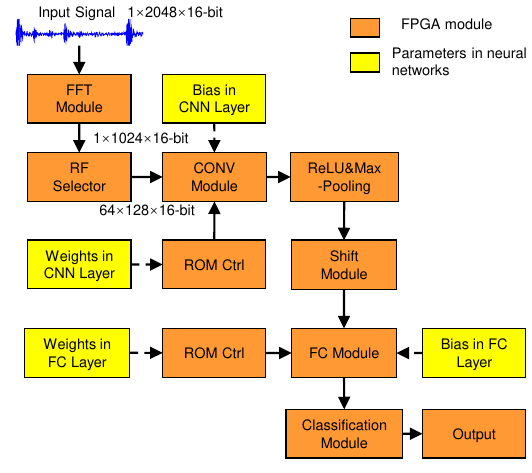}
    \caption{The overall diagram for deploying the BearingPGA-Net into FPGA.}
    \label{fig:implestru}
    \vspace{-0.3cm}
\end{figure}

\underline{Receiving filter (RF) selector}: It splits the input signal into 128 segments of 64 points with a stride of 8, which is equivalent to sliding a convolutional kernel of $1\times64$ with a stride of 8 over the input signal.

\underline{ROM control (Ctrl)}: It retrieves weight parameters in ROM for the convolutional and fully-connected layers. For the convolutional layer, it reads 256 convolutional weight parameters (4 $\times$ 64 kernels) once a time and 10 weight parameters per cycle to update the input of the multiplication-accumulation module for the fully-connected layer. In addition, the FPGA directly stores bias parameters (4 for the convolutional layer and 10 for the fully-connected layer) in registers.

\underline{Shift module}: It shifts the position of the decimal point, which converts the bit-width of integers and decimals from (2,13) to (7,8) to adjust the fixed-point numbers in the downstream fully-connected layer.

\underline{Classification module}: It selects the maximum value among 10 outputs of the fully-connected layer as the failure category of the bearing. Softmax is waived in FPGA deployment.








\textbf{3) BearingPGA-Net accelerator.} CNN accelerator designs can be generally classified into two groups: computing acceleration and memory system optimization \cite{qiu2016going}. Since our BearingPGA-Net consists of only one convolutional layer and one fully-connected layer, memory management does not require additional optimization. We focus on optimizing the FPGA acceleration of each layer in BearingPGA-Net.

\begin{figure}[h]
    \centering
    \includegraphics[width=\linewidth]{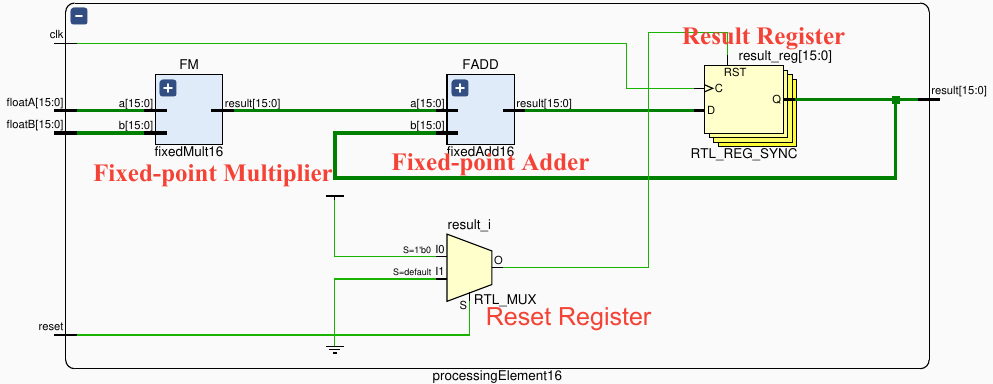}
    \caption{The circuit diagram of a multiplication-accumulation unit, which includes a fixed-point multiplier, a fixed-point adder, a reset register, and a result register.}
    \label{fig:circuit}
    \vspace{-0.3cm}
\end{figure}

\underline{Multiplication-accumulation unit.} As FPGA registers cannot directly perform matrix calculations, we design the multiplication-accumulation unit beforehand. The logic circuit diagram of the multiplication-accumulation unit is shown in Fig.~\ref{fig:circuit}, where the accumulation operation is executed in $N$ cycles, and the result register is used to store the temporary results in each cycle. The expression of the multiplication-accumulation unit is
\begin{equation}
    y = {\textstyle \sum}_{i}^N(p_i \times q_i),
\end{equation}
where $p_i$ and $q_i$ are 16-bit fixed-point numbers.

\begin{figure}[h]
    \centering
    \includegraphics[width=0.75\linewidth]{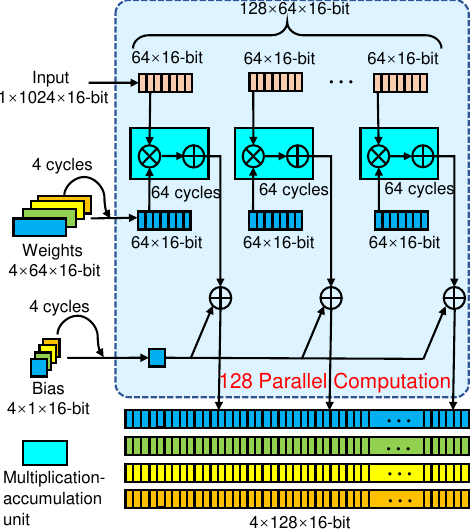}
    \caption{The implementation diagram of the convolutional layer.}
    \label{fig:Convstructure}
    \vspace{-0.6cm}
\end{figure}

\underline{Convolutional layer.} Fig. \ref{fig:Convstructure} illustrates the implementation of the convolution layer. The signal after passing through the RF selector is divided into segments of $128 \times 64$. Subsequently, the multiplication-accumulation unit uses weights $w \in \mathbb{R}^{1 \times 64}$ stored in the ROM to multiply the signal $s_i \in \mathbb{R}^{1 \times 64}$. Each multiplication-accumulation unit runs for 64 cycles and adds the bias $b$. The equation for this operation is $z_i = \sum_{j=1}^{64}(s_{i,j} \times w_{j})+b$.
Moreover, hardware acceleration is achieved using 128 parallel multiplication-accumulation units. This enables each convolutional kernel to complete the full convolution in just 64 cycles. The parallel output is merged into $\boldsymbol{z}=\{z_1,z_2,\cdots, z_{128}\}$. With 4 convolutional kernels, the layer requires 4 external cycles, totaling 256 cycles. The speedup is 128-fold relative to conventional implementations.


\underline{ReLU and max-pooling layer}.
ReLU activation and max-pooling are back-to-back in the BearingPGA-Net. Also, both ReLU and max-pooling functions attempt to find the maximum value between two numbers. Therefore, we fuse them into a single module to save the inference time. Suppose that the input signal is $\boldsymbol{x} \in \mathbb{R}^{1 \times n}$, the ReLU activation function is
\begin{equation}
    p_i = \max(0, x_i), \ i=1,2,\cdots, n,
\end{equation}
followed by the max-pooling layer, where the window size is $k$ and the stride is $s$. The output of this layer is 
\begin{equation}
    q_i = \max (\{p_j\}_{j=i\times s}^{i\times s + k -1}), \ i=1,2,\cdots, \lfloor \frac{n-k}{s}  \rfloor +1,
\end{equation}
where $\lfloor \cdot \rfloor$ is the floor function to ensure the last pooling window is fully contained within the signal length.

We devise a strategy to reduce the amount of computation as Algorithm \ref{alg:max} shows. Two 16-bit numbers, denoted as $x_1$ and $x_2$, are firstly compared in terms of their symbol bit. If both numbers are smaller than 0, the output is set to 0, thus executing the ReLU function directly. On the other hand, if the numbers have opposite signs, the negative number is killed. These operations can save one maximum comparison. Numerical comparisons are only performed when both numbers are positive, in which case the complete ReLU and max-pooling operations are executed. By adopting our method, we are able to save approximately $2/3$ of LUT resources.

\begin{algorithm}
\caption{ReLU and Max-pooling}\label{alg:max}
\begin{algorithmic}[1]
\Require $x_1, x_2$ ($x[15]$ symbol bit, $x[14:0]$ number bits)
\If{$x_1[15]>0$ AND $x_2[15]<0$}
    \State \Return $x_1$
\ElsIf{$x_1[15]<0$ AND $x_2[15]>0$}
    \State \Return $x_2$
\ElsIf{$x_1[15]<0$ AND $x_2[15]<0$}
    \State \Return $0$
\ElsIf{$x_1[15]>0$ AND $x_2[15]>0$}
    \If{$x_1[14:0] >= x_2[14:0]$} 
        \State \Return $x_1$
    \ElsIf{$x_1[14:0] < x_2[14:0]$}
        \State \Return $x_2$
    \EndIf
\EndIf
\end{algorithmic}
\end{algorithm}

\begin{figure}[h]
    \centering
    \includegraphics[width=0.75\linewidth]{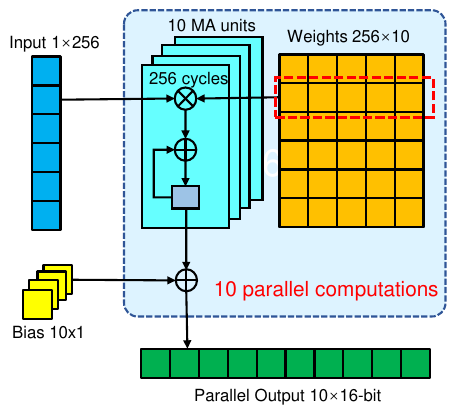}
    \caption{The implementation diagram of the fully-connected layer.}
    \label{fig:fcstructure}
    \vspace{-0.6cm}
\end{figure}

\underline{Fully-connected layer}. The fully-connected layer establishes connections between the 256 outputs of the ReLU and max-pooling fusion layers and the final 10 outputs. Unlike on computers, the fully-connected layer in FPGAs consumes fewer resources, as there is no need to slide windows. For the fully-connected layer, we reuse multiplication-accumulation units. Initially, the size of weights in the fully-connected layer is $256 \times 10$. To simplify the computations, we divide this matrix into 256 segments, each consisting of 10 weights. Consequently, 10 multiplication-accumulation units perform parallel computation over 256 cycles to generate the output. The implementation for this design is illustrated in Fig. \ref{fig:fcstructure}.

In summary, our FPGA accelerator offers two main advantages: i) Parallelism: multiplication-accumulation units are simultaneously calculated, which greatly boosts computational efficiency. ii) Module reuse: due to the constraint of limited FPGA computing resources that cannot parallelize 1024 units concurrently, multiplication-accumulation units are reused to strike a balance between resources and speed.

\section{Experiments}

\subsection{Datasets Descriptions}

\textbf{1) CWRU dataset.} This widely-used dataset is curated by Case Western Reserve University Bearing Data Center, which consists of two deep groove ball bearings mounted on the fan-end (FE) and drive-end (DE) of the electric motor. Electro-discharge machining is used to inject single-point defects with diameters of 7, 14, and 21 mils into the outer race, inner race, and ball of both bearings, respectively. As a result, there are ten categories in this dataset, including nine types of faulty bearings and one healthy bearing. Specifically, the motor shaft is subjected to four levels of load (0HP, 1HP, 2HP, 3HP, where HP denotes horsepower), which slightly affects the motor speed (1797 r/min, 1772 r/min, 1750 r/min, 1730 r/min). Vibration data are collected at 12 kHz and 48 kHz, respectively. In this study, we analyze the vibration signals collected at 12 kHz on the DE side of the motor.

\begin{table}[htbp]
\caption{Ten healthy statuses in our HIT dataset. OR and IR denote that the faults appear in the outer race and inner race, respectively.}
\centering
\begin{tabular}{|l|l|l|l|}
\hline 
Label & Faulty Mode        & Label & Faulty Mode  \\ \hline 
1    & Health                                    & 6    & OR (Moderate)  \\ \hline
2    & Ball cracking (Minor)       & 7    & OR (Severe)   \\ \hline
3    & Ball cracking (Moderate)     & 8    & IR (Minor)   \\ \hline
4    & Ball cracking (Severe)       & 9    & IR (Moderate) \\ \hline
5    & OR cracking (Minor) & 10   & IR (Severe)   \\ \hline
\end{tabular}
\label{tab:class}
\end{table}

\begin{figure}[h]
\vspace{-0.2cm}
    \centering
\includegraphics[width=0.8\linewidth]{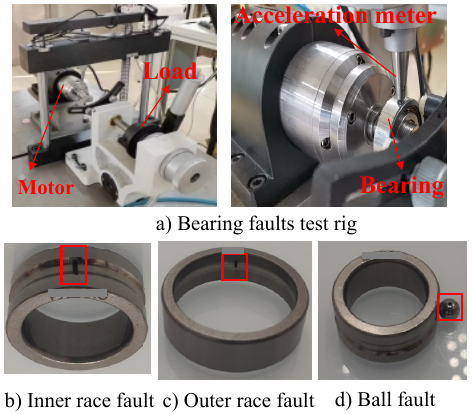}
    \caption{The test rig of faulty bearings to collect data.}
    \label{fig:HITdataset}
    \vspace{-0.6cm}
\end{figure}

\textbf{2) HIT dataset.} The bearing fault test is conducted in the MIIT Key Laboratory of Aerospace Bearing Technology and Equipment at Harbin Institute of Technology (HIT). Fig. \ref{fig:HITdataset} shows the bearing test rig and the faulty bearings used in the experiment. We utilize HC7003 angular contact ball bearings, which are designed for high-speed rotating machines. The accelerometer is directly attached to each bearing to collect vibration signals of the bearings. Similar to the CWRU dataset, we injected defects at the outer race (OR), inner race (IR), and ball at three severity levels (minor, moderate, severe). Tab.~ \ref{tab:class} presents ten healthy statuses of bearings in our dataset. For the test, a constant motor speed of 1800 r/min is set, and vibration signals are acquired using the NI USB-6002 device at a sampling rate of 12 kHz. Each bearing vibration is recorded for 47 seconds, resulting in 561,152 data points per category. Our dataset is more challenging than the CWRU dataset because the bearing faults in our dataset are cracks of the same size but varying depths whose vibration signals between different faults exhibit more similarity.

\subsection{Experimental Configurations}
\textbf{1) Data preprocessing.} First, the raw long signal is divided into segments of 2,048 points and standardized. Next, the Fast Fourier Transform (FFT) is applied to convert the signal from the time domain to the frequency domain. Previous research has experimentally demonstrated that employing signal processing techniques such as FFT and wavelet transform, can enhance the performance of shallow neural networks~\cite{zhao2020deep}. Because the characteristics of non-stationary vibration signals are more pronounced in the frequency or time-frequency domain than the time domain, this compensates for the constrained feature extraction capability of shallow networks.

Specifically, the starting point of the segment of 2,048 points is randomly chosen, and the stride is set to 28 to resample the raw signal. All classes of signals are sampled 1,000 times, resulting in a total of 10,000 samples. Then, the dataset is randomly divided into training, validation, and testing sets with a ratio of 2:1:1. Next, the Gaussian noise is added to the signals to simulate noise in the industrial fields and also evaluate different models' performance in noisy experiments. The signal-to-noise ratio (SNR) is calculated as $\text{SNR} =10\log 10(P_s/P_n)$, where $P_s$ and $P_n$ are the average power of the signal and noise, respectively. Lastly, the signals are standardized using z-score standardization. Notably, each signal $\boldsymbol{x}$ is transformed as $\boldsymbol{x}'=(\boldsymbol{x}-\mu)/\sigma$, where $\mu$ and $\sigma$ are the mean and standard deviation of the signal $\boldsymbol{x}$, respectively. For the CWRU dataset, the SNR ranges from -6dB to 2dB in 2dB intervals. For our dataset, the SNR ranges from 0dB to 8dB in 2dB intervals. 

\begin{figure}[h]
    \vspace{-0.5cm}
    \centering
    \includegraphics{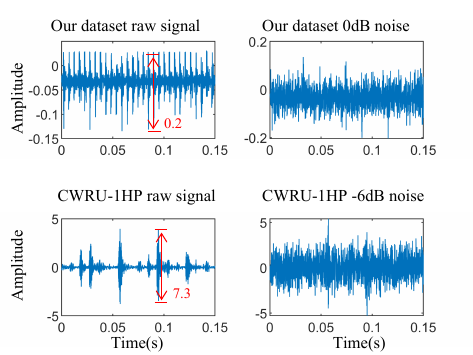}
    \caption{Comparison of two datasets under noise.}
    \label{fig:sig}
    \vspace{-0.2cm}
\end{figure}

As shown in Fig.~\ref{fig:sig}, the signal amplitude in our dataset is lower than in the CWRU dataset. Therefore, even relatively low-intensity noise (0dB SNR) can overshadow the signal.

\textbf{2) Baselines.} We compare our method against four state-of-the-art lightweight baselines: wide deep convolutional neural networks (WDCNN)~\cite{zhang2017new}, lightweight efficient feature extraction networks (LEFE-Net)~\cite{fang2021lefe}, lightweight transformers with convolutional embeddings and linear self-attention (CLFormer)~\cite{fang2021clformer}, and knowledge distillation-based student convolutional neural networks (KDSCNN)~\cite{ji2022neural}. All these models are published in flagship journals of this field in recent years. As summarized in Tab.~\ref{tab:model_para}, our model enjoys the smallest model size, relatively low computational complexity, and the second shortest inference time compared to its competitors. Although our model has slightly higher computational complexity and inference time than KDSCNN due to the introduced FFT, it has a substantially smaller number of parameters (2.83K vs 5.89K), while the model size is the first priority for deployment on FPGAs.

\begin{table}[htbp]
\centering
\caption{The properties of compared models. \#FLOPs denotes floating point operations. Time is the elapsed time to infer 1000 samples on an NVIDIA Geforce GTX 3080 GPU.}
\begin{tabular}{|l|c|c|c|}
\hline
Method    & \#Parameters  & \#FLOPs & Inference time \\ \hline
LEFE-Net & 73.95K       & 14.7M    & 22.4540s       \\ \hline
WDCNN     & 66.79K      & 1.61M    & 1.6327s        \\ \hline
CLFormer  & 4.98K        & 0.63M    & 0.7697s        \\ \hline
KDSCNN  & 5.89K        & 70.66K   & 0.2219s        \\ \hline
BearingPGA-Net      & 2.83K        & 78.34K   & 0.6431s        \\ \hline
\end{tabular}
\label{tab:model_para}
\end{table}

\textbf{3) Implementation settings.} All experiments are conducted in Windows 10 with Intel(R) Core(TM) 11th Gen i5-1135G7 at 2.4GHz CPU and one NVIDIA RTX 3080 8GB GPU. Our code is written in Python 3.10 with PyTorch 2.1.0.

For all compared models, we use stochastic gradient descent (SGD) \cite{bottou2012stochastic} as the optimizer with momentum=0.9 and Cosine Annealing LR \cite{loshchilov2016sgdr} as the learning rate scheduler. The training epochs are set to 75, and we use grid search to find the best hyperparameters. $[32, 64, 128, 256]$ is for the batch size, and $[10^{-4}, 3\times10^{-4}, 10^{-3}, 3\times10^{-3}, 0.01, 0.03, 0.1, 0.3]$ is for the learning rate. In KD-based methods, we search $T$ from $[1,2,3,4,5,6,7,8,9]$, and $\alpha$ from $[0.1, 0.2, 0.3, 0.4, 0.5, 0.6, 0.7, 0.8, 0.9]$, while in our model we have to additionally search $\beta$ and $\gamma$. We configure $[0, 0.2, 0.5, 1, 2, 4]$ for $\beta$ and $[1, 2, 4, 8, 10]$ for $\gamma$, respectively.

\textbf{4) Evaluation metrics.} We use the F1 score to validate the performance of the proposed method. The metric is defined as 
\begin{equation}
\nonumber
\mathrm{F1} \ \mathrm{score} = \frac{1}{\mathrm{C}}  {\textstyle \sum}^{\mathrm{C}}_{i=1} \frac{2\mathrm{TP}_i}{2\mathrm{TP}_i + \mathrm{FP}_i + \mathrm{FN}_i }, 
\end{equation}
where $\mathrm{TP}_i$, $\mathrm{FP}_i$, and $\mathrm{FN}_i$ are the numbers of true positives, false positives, and false negatives for the $i$-th class. $\mathrm{C}$ denotes the number of classes. All results are the average of ten runs.

\begin{figure}[htbp]
\vspace{-0.3cm}
    \centering
    \includegraphics[width=\linewidth]{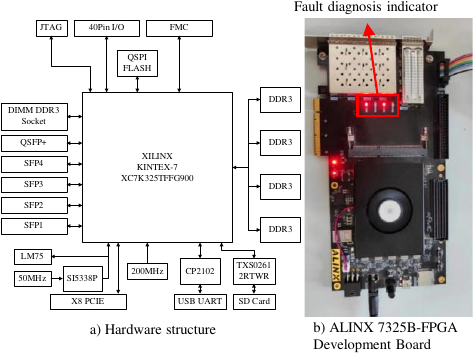}
    \caption{The hardware architecture and physical diagram of the FPGA where BearingPGA is deployed.}
    \vspace{-0.3cm}
    \label{fig:FPGAboard}
\end{figure}

\textbf{5) FPGA description.} We utilize the ALINX 7325B-FPGA as the hardware deployment board, as shown in Fig.~\ref{fig:FPGAboard}. This board incorporates the Kintex-7 XC7K325T FPGA chip, which is widely used in the industrial field due to its optimal balance between price and performance-per-watt. The clock frequency is set to 100MHz, and we utilize four LEDs as indicators of the fault diagnosis result, where ten categories are encoded as 4-bit binary numbers. The specifications of the FPGA chip are summarized in Tab.~\ref{tab:FPGAresource}.

\begin{table}[htbp]
\centering
\caption{Specifications of Kintex-7 XC7K325T}
\begin{tabular}{|l|c|}
\hline
Name              & Parameter        \\ \hline
Logic Cells       & 326,080          \\
Slices            & 50,950           \\
CLB flip-flops    & 407,600          \\
Block RAM(KB)     & 16,020           \\
DSP48 Slices      & 840              \\
PCIE Gen2         & 1                \\
XADC              & 12bits, 1Mbps AD \\
GTP Transceiver   & 16 per chip, 12.5Gb/s     \\
Speed Grade       & -2               \\ \hline
\end{tabular}
\label{tab:FPGAresource}
\vspace{-0.3cm}
\end{table}

\begin{table*}[h]
\centering
\caption{The F1 scores (\%) of all models on the CWRU dataset. The bold-faced number is the best, and the underline number is the second best.}
\begin{tabular}{|ll|cccccc|c|}
\hline
                          & Model       & -6dB                  & -4dB                  & -2dB                   & 0dB                    & 2dB                    & Clean                  & Average           \\ \hline
\multirow{5}{*}{CWRU-0HP} & LEFE-Net & 86.62$\pm$3.99          & 94.34$\pm$2.42          & 98.69$\pm$1.33           & 98.89$\pm$1.14           & 98.96$\pm$1.05           & 99.87$\pm$0.10           & 98.15          \\
                          & WDCNN     & 95.44$\pm$2.21          & \textbf{98.33$\pm$1.11} & {\ul 98.52$\pm$1.05}     & {\ul 99.04$\pm$0.63}     & 99.00$\pm$0.32           & \textbf{100.00$\pm$0.00} & {\ul 98.39}    \\
                          & CLFormer  & 85.27$\pm$3.71          & 89.84$\pm$3.13          & 96.11$\pm$3.31           & 98.75$\pm$1.32           & {\ul 99.32$\pm$0.23}     & 99.46$\pm$0.13           & 94.79          \\
                          & KDSCNN  & 79.95$\pm$4.35          & 84.90$\pm$4.01           & 87.57$\pm$3.53           & 93.55$\pm$2.36           & 96.76$\pm$2.41           & 98.25$\pm$1.70           & 90.16          \\
                          & BearingPGA-Net      & \textbf{97.20+2.16}  & {\ul 98.23$\pm$2.07}    & \textbf{99.20$\pm$0.01}   & \textbf{99.89$\pm$0.11}  & \textbf{99.98$\pm$0.01}  & \textbf{100.00$\pm$0.00} & \textbf{99.08} \\ \hline
\multirow{5}{*}{CWRU-1HP} & LEFE-Net & 75.95$\pm$3.41          & 82.42$\pm$2.33          & 92.40$\pm$2.01            & 97.62$\pm$2.36           & {\ul 97.51$\pm$2.43}     & {\ul 99.24$\pm$0.45}     & 90.86          \\
                          & WDCNN     & {\ul 92.24$\pm$2.35}    & {\ul 93.74$\pm$2.56}    & \textbf{97.88$\pm$2.10}  & {\ul 97.64$\pm$2.07}     & 97.48$\pm$2.33           & 98.47$\pm$1.29           & {\ul 96.24}    \\
                          & CLFormer  & 65.74$\pm$3.75          & 70.94$\pm$3.11          & 88.22$\pm$3.14           & 90.16$\pm$1.65           & 95.90$\pm$1.01           & 99.02$\pm$0.89           & 85.00          \\
                          & KDSCNN  & 85.50$\pm$3.36           & 87.76$\pm$3.51          & 89.43$\pm$2.27           & 91.99$\pm$2.33           & 96.52$\pm$1.11           & 97.65$\pm$1.38           & 91.48          \\
                          & BearingPGA-Net      & \textbf{96.85$\pm$1.11} & \textbf{97.56$\pm$1.14} & {\ul 97.75$\pm$1.71}     & \textbf{97.96$\pm$2.64}  & \textbf{98.08$\pm$1.85}  & \textbf{99.97$\pm$0.01}  & \textbf{98.03} \\ \hline
\multirow{5}{*}{CWRU-2HP} & LEFE-Net & 91.56$\pm$1.14          & 97.68$\pm$2.31          & {\ul 99.96$\pm$0.04}     & 99.84$\pm$0.19           & 99.94$\pm$0.06           & 99.93$\pm$0.02           & 98.15          \\
                          & WDCNN     & \textbf{99.40$\pm$0.01}  & \textbf{99.99$\pm$0.01} & \textbf{100.00$\pm$0.00} & \textbf{100.00$\pm$0.00} & \textbf{100.00$\pm$0.00} & \textbf{100.00$\pm$0.00} & \textbf{99.90} \\
                          & CLFormer  & 88.47$\pm$3.85          & 89.84$\pm$3.98          & 97.31$\pm$1.59           & 99.84$\pm$0.05           & 99.85$\pm$0.14           & 99.92$\pm$0.03           & 95.87          \\
                          & KDSCNN  & 94.68$\pm$2.92          & 97.56$\pm$1.41          & 98.23$\pm$1.63           & 98.60$\pm$1.04           & 98.98$\pm$1.01           & 99.10$\pm$0.84           & 97.86          \\
                          & BearingPGA-Net      & {\ul 98.66$\pm$1.01}    & {\ul 99.56$\pm$0.33}    & 99.77$\pm$0.14           & {\ul 99.94$\pm$0.01}     & \textbf{100.00$\pm$0.00} & \textbf{100.00$\pm$0.00} & {\ul 99.66}    \\ \hline
\multirow{5}{*}{CWRU-3HP} & LEFE-Net & 89.65$\pm$3.39          & 95.13$\pm$1.15          & 96.87$\pm$3.11           & 99.56$\pm$0.41           & \textbf{99.99$\pm$0.01}  & 99.97$\pm$0.39           & 96.86          \\
                          & WDCNN     & \textbf{95.83$\pm$1.31} & {\ul 98.65$\pm$1.31}    & \textbf{99.84$\pm$0.17}  & \textbf{99.96$\pm$0.02}  & \textbf{99.99$\pm$0.01}  & \textbf{100.00$\pm$0.00} & \textbf{99.05} \\
                          & CLFormer  & 87.59$\pm$3.13          & 89.53$\pm$3.11          & 97.19$\pm$1.11           & 98.92$\pm$0.53           & 99.28$\pm$0.27           & 99.92$\pm$0.01           & 95.41          \\
                          & KDSCNN  & 83.15$\pm$3.95          & 86.58$\pm$2.99          & 95.95$\pm$1.33           & 96.54$\pm$1.63           & 98.17$\pm$1.64           & 99.71$\pm$0.14           & 93.35          \\
                          & BearingPGA-Net      & {\ul 95.47$\pm$1.39}    & \textbf{98.93$\pm$1.03} & {\ul 99.15$\pm$0.61}     & {\ul 99.68$\pm$0.31}     & 99.78$\pm$0.22           & \textbf{100.00$\pm$0.00} & {\ul 98.84}    \\ \hline
\end{tabular}
\label{tab:CWRU_results}
\end{table*}

\subsection{Classification Results}

Here, we compare the performance of the student model (BearingPGA-Net) with its competitors.

\textbf{1) Results on the CWRU and HIT datasets.}
On the CWRU dataset, as shown in Tab. \ref{tab:CWRU_results}, BearingPGA-Net achieves the top F1 scores in 0HP and 1HP conditions and the second-best scores in 2HP and 3HP conditions. Moreover, our method maintains over 95\% F1 score at the -6dB noise level, whereas the KDSCNN performs the worst on the CWRU-0HP dataset at -6dB noise (79.95\%). Although WDCNN performs comparably, its number of parameters is approximately 22 times larger than ours. These results demonstrate BearingPGA's strong diagnosis performance despite its small parameter size.

On the HIT dataset, as Tab. \ref{tab:hitresults} shows, our method achieves the highest average F1 score in noisy scenarios. Although LEFE-Net slightly outperforms ours on clean signals, the difference is only 1\%, and the LEFE-Net has a substantially larger model size. Furthermore, the performance of other models (CLFormer: 76.37\%, KDSCNN: 74.98\%) falls behind ours (83.30\%) by a large margin. Overall, our model is the best performer on this dataset.

\begin{table}[h]
\caption{The F1 score (\%) of all compared methods on the HIT dataset. Where N/A denotes the raw signal without noise, the bold-face number is the best.}
\centering
\begin{tabular}{|l|c|c|c|c|c|}
\hline
SNR     & LEFE-Net        & WDCNN   & CLFormer & KDSCNN       & Ours                                    \\ \hline
0       & 58.71        & 58.53 & 54.61  & 64.78                         & \textbf{70.51}                        \\
2       & 71.27          & 68.92 & 64.88  & 70.02                        & \textbf{72.64}                        \\
4       & 79.30          & 78.96 & 77.26  & 73.64                        & \textbf{82.46}                        \\
6       & 83.31          & 84.60 & 78.38  & {76.62} & {\textbf{85.53}} \\
8       & 86.28          & 90.92 & 89.91  & {79.10} & {\textbf{91.54}} \\
Clean     & \textbf{98.96} & 94.35 & 93.15  & {85.72} & {97.39}          \\
Average & 79.64          & 79.38 & 76.37  & 74.98                        & \textbf{83.35}                        \\ \hline
\end{tabular}
\label{tab:hitresults}
\end{table}

\textbf{2) Model compression results.}
We compare the compression performance of DKD. Firstly, Tab.~\ref{tab:TvSproper} shows the properties of teacher and student models. DKD effectively compresses the teacher model, reducing model parameters by 17 times, FLOPs by 10.6 times, and inference time by 2.45 times. Secondly, Tab.~\ref{tab:TvS} compares the performance of the models. Clearly, DKD enhances the performance of student model, though still slightly below the teacher (with an average F1 score approximately 1\% lower). Comparing the student model with and without DKD, distillation provides considerable improvements, especially in noisy environments. On the CWRU-3HP, DKD improves the average F1 score by 3.7\%. Notably, DKD admits 8.23\% and 8.46\% gains at -6dB on CWRU-2HP and 0dB on the HIT dataset, respectively. These results demonstrate DKD can successfully transfer knowledge to shallow models, despite substantial compression.

\begin{table}[h]
\centering
\caption{The properties of teacher and student models. \#FLOPs denotes floating point operations, lower is better. Time is the elapsed time to infer 1000 samples}
\begin{tabular}{|l|c|c|c|}
\hline
Model   & \#Parameters  & \#FLOPS & Inference Time    \\ \hline
Teacher & 50.09K      & 0.83M   & 1.5703s \\
Student & 2.83K       & 78.34K  & 0.6431s \\ \hline
\end{tabular}
\label{tab:TvSproper}
\vspace{-0.4cm}
\end{table}

\begin{table}[h]
\centering
\caption{The average F1 score (\%) of teacher and student models and student models without decoupled knowledge distillation.}
\begin{tabular}{|l|c|c|c|c|}
\hline
Dataset                   & SNR  & Teacher & Student & Student w/o KD \\ \hline
\multirow{6}{*}{CWRU-0HP} & -6   & 98.57   & 97.20   & 93.20                         \\
                          & -4   & 99.69   & 98.23   & 99.04                         \\
                          & -2   & 99.79   & 99.20   & 99.07                         \\
                          & 0    & 99.53   & 99.89   & 99.67                         \\
                          & 2    & 100.00  & 99.98   & 99.64                         \\
                          & Avg. & 99.52   & 98.83   & 98.12                         \\ \hline
\multirow{6}{*}{CWRU-1HP} & -6   & 97.21   & 96.85   & 88.62                         \\
                          & -4   & 97.92   & 97.56   & 92.62                         \\
                          & -2   & 97.73   & 97.75   & 95.77                         \\
                          & 0    & 98.20   & 97.96   & 96.53                         \\
                          & 2    & 99.41   & 98.08   & 96.78                         \\
                          & Avg. & 98.09   & 97.64   & 94.06                         \\ \hline
\multirow{6}{*}{CWRU-2HP} & -6   & 99.42   & 98.66   & 96.45                         \\
                          & -4   & 99.89   & 99.56   & 98.85                         \\
                          & -2   & 99.64   & 99.77   & 99.85                         \\
                          & 0    & 100.00  & 99.94   & 99.78                         \\
                          & 2    & 100.00  & 100.00  & 99.99                         \\
                          & Avg. & 99.79   & 99.59   & 98.98                         \\ \hline
\multirow{6}{*}{CWRU-3HP} & -6   & 97.97   & 95.47   & 91.62                         \\
                          & -4   & 98.98   & 98.93   & 94.22                         \\
                          & -2   & 99.62   & 99.15   & 95.28                         \\
                          & 0    & 99.72   & 99.68   & 96.71                         \\
                          & 2    & 100.00  & 99.78   & 97.06                         \\
                          & Avg. & 99.26   & 98.60   & 94.98                         \\ \hline
\multirow{6}{*}{HIT}     & 0    & 71.36   & 70.51   & 62.05                         \\
                          & 2    & 73.77   & 72.64   & 72.68                         \\
                          & 4    & 84.15   & 82.46   & 80.92                         \\
                          & 6    & 85.66   & 85.54   & 85.10                         \\
                          & 8    & 93.12   & 91.54   & 90.67                         \\
                          & Avg. & 81.61   & 80.54   & 78.28                         \\ \hline
\end{tabular}
\label{tab:TvS}
\vspace{-0.2cm}
\end{table}

\underline{Ablation study.}
The ablation experiments investigate the roles of TCKD and NCKD in decoupled knowledge distillation. The experiments are divided into classical KD, TCKD performed independently, NCKD performed independently, and DKD. Note that DKD involves two loss functions, with $\beta=4$ and $\gamma=1$, while for the independent TCKD or NCKD, one hyperparameter is assigned a value of 1 and the other is set to 0.

As shown in Tab. \ref{tab:ablation}, the performance of classical KD is superior to that of two independent loss functions, and the F1 score of independent TCKD is comparable to that of classical KD. In contrast, independent NCKD performs the worst. This phenomenon suggests that TCKD plays a major role in the knowledge distillation loss function. Moreover, the DKD loss demonstrates the best results. It is noteworthy that the weight of TCKD is set to four times that of NCKD, indicating that focusing the attention of neural networks more on TCKD can further unleash the potential of logit knowledge distillation.

\begin{table}[h]
\centering
\caption{The F1 score of different KD functions.}
\scalebox{0.8}{
\begin{tabular}{@{}|cc|ccccc|@{}}
\hline
TCKD & NCKD & 0HP(-6dB) & 1HP(-6dB) & 2HP(-6dB) & 3HP(-6dB) & HIT(8dB) \\ \hline
\textemdash    & \textemdash    & 94.32     & 91.00     & 97.62     & 92.14     & 82.15     \\
\Checkmark    & \XSolidBrush    & 94.31     & 90.22     & 97.57     & 92.11     & 82.15     \\
\XSolidBrush    & \Checkmark    & 88.80      & 91.38     & 96.03     & 89.36     & 75.81     \\
\Checkmark    & \Checkmark    & \textbf{97.20}     & \textbf{96.85}     & \textbf{98.66}     & \textbf{95.47}     & \textbf{93.12}     \\ \hline
\end{tabular}}
\label{tab:ablation}
\vspace{-0.4cm}
\end{table}

\underline{Hyperparameter sensitivity.}
Experiments are conducted to verify the hyperparameters in DKD, namely $T$, $\alpha$, $\beta$, and $\gamma$. In this experiment, we set the learning rate to 0.1 and the batch size to 64. DKD's default hyperparameters are $T=2.5, \alpha=0.2, \beta=4, \gamma=1$. We then test the performance on the CWRU-0HP dataset at -6dB noise by changing one of these hyperparameters.

\begin{table}[h]
\centering
\caption{The hyperparameter sensitivity in DKD on the CWRU-0HP dataset with -6dB noise.}
\begin{tabular}{@{}|l|ccccc|@{}}
\hline
$T$     & 1.5   & 2     & 2.5   & 3     & 3.5   \\ \hline
F1(\%)      & 95.28 & 94.85 & 95.20 &  \textbf{96.19} & 96.18 \\ \hline
$\alpha$ & 0.1   & 0.2   & 0.3   & 0.4   & 0.5   \\ \hline
F1(\%)      & 92.49 & \textbf{96.19} & 32.13 & 32.25 & 32.19 \\ \hline
$\beta$  & 0.2   & 0.5   & 1     & 2     & 4     \\ \hline
F1(\%)     & 89.49 & 91.08 & 92.84 & 95.07 & \textbf{96.20} \\ \hline
$\gamma$ & 0.2   & 0.5   & 1     & 2     & 4     \\ \hline
F1(\%)      & \textbf{96.41} & 95.68 & 96.19 & 96.30  & 32.09 \\ \hline
\end{tabular}
\label{tab:hyperparam}
\vspace{-0.4cm}
\end{table}

The results are presented in Tab.~ \ref{tab:hyperparam}. Firstly, the parameter of temperature ($T$) does not appear to be highly influential. While a larger value of $T$ leads to slightly better performance, the improvement is only about 1\%. Secondly, the analysis shows that all three scale parameters have a significant impact. In particular, setting $\alpha$ to 0.2 yields superior results. Once $\alpha$ surpasses 0.3, there is a noticeable decline in performance, with approximately 64\% drop in F1 score. Unlike $\alpha$, $\beta$ exhibits less sensitivity, but increasing its value still leads to improved results. Conversely, a lower value (below 2) is recommended for the parameter $\gamma$.

The aforementioned phenomenon exemplifies the characteristics of DKD. Regarding the parameter $\alpha$, it is utilized to regulate both $\mathcal{L}_{CE}$ and $\mathcal{L}_{KL}$ loss. A lower value of $\alpha$ implies that the student network predominantly learns from the ground truth, while any knowledge transmitted by the teacher model serves merely as supplementary information. Next, $\beta$ and $\gamma$ are employed to control the TCKD and NCKD losses, respectively. Consistent with the findings of the ablation experiments, a larger value for $\beta$ and a smaller value for $\gamma$ yield optimal results, as they compel the model to effectively harness the knowledge pertaining to the target classes.

\textbf{3) Classification results on FPGA.}
We utilize the HIT dataset as the input to compare the performance of the 32-bit float PyTorch and the 16-bit fixed-point FPGA implementation. Tab.~\ref{tab:fpgaresults} reveals that the quantized model demonstrates a lower-than-0.4\% performance drop in terms of F1, Recall, and Precision scores relative to the original model. Additionally, the parameters in the quantized model are reduced by more than half. These findings highlight the effectiveness of the designed quantization method and convolution architecture for FPGA deployment.

\begin{figure}[H]
\vspace{-0.4cm}
    \centering
    \includegraphics{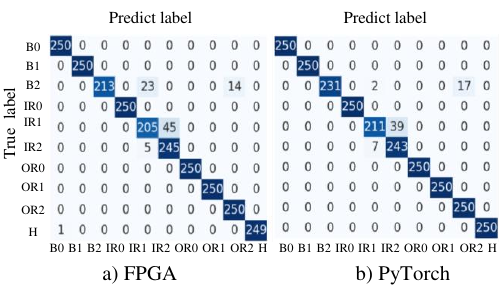}
    \caption{The confusion matrices of results produced by models before and after quantization on the HIT dataset, where B, IR, OR denote faults at the ball, the inner race, and the outer race, respectively, and H denotes the healthy bearing.}
\vspace{-0.3cm}
    \label{fig:confusion}
\end{figure}

\begin{table}[h]
\vspace{-0.2cm}
\centering
\caption{Comparsion of 32-bit Pytorch and 16-bit FPGA models on HIT dataset.}
\begin{tabular}{|l|cccc|}
\hline
    &\#Parameters    & F1 (\%)       & Recall (\%)          & Precision (\%)       \\ \hline
PyTorch &2.83K & 97.39          & 97.40          & 97.57          \\
\hline
FPGA    &1.42K & {97.12} & {97.34} & {97.12} \\ \hline
\end{tabular}
\vspace{-0.2cm}
\label{tab:fpgaresults}
\end{table}
Furthermore, the confusion matrices between the FPGA and PyTorch models are compared. As illustrated in Fig. \ref{fig:confusion}, the FPGA model exhibits $>18$ errors in classifying ball faults (B0-B2), $<4$ errors for inner race faults (IR0-IR2), and 1 additional error in healthy bearings, compared to PyTorch results. The BearingPGA-Net when deployed on FPGA only generates a total of 24 extra errors compared to the original version across 2500 samples.

\begin{figure}[h]
    \centering
    \includegraphics[width=0.8\linewidth]{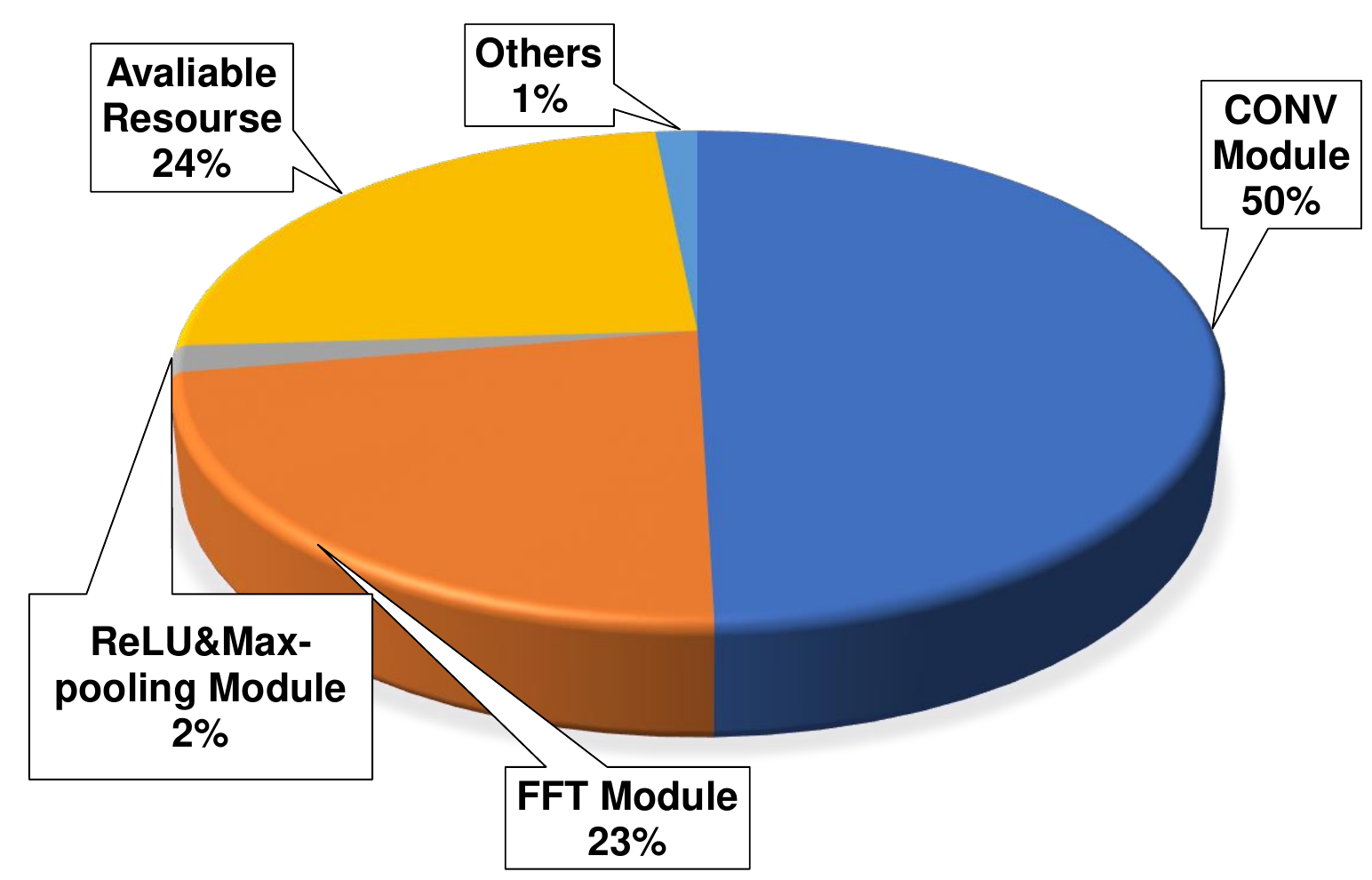}
    \caption{The LUT consumption with respect to each module.}
    \label{fig:pie}
\end{figure}

\textbf{4) Resource analysis.}
The resource usage for the entire network is summarized in Tab. \ref{tab:resource}. The LUT resource and block RAM (BRAM) resource exhibit utilization rates of 74.40\% and 58.20\%, respectively, indicating that BearingPGA-Net makes good use of the FPGA's computational and memory resources. Next, we analyze the resource consumption specifically within the LUT, which is the key computational resource in an FPGA. As illustrated in Fig. \ref{fig:pie}, the convolution and FFT modules account for the majority of resources in the LUT, constituting approximately 50\% and 23\%, respectively. This observation underscores the limited computational resources available on a Kintex-7 FPGA, \textit{i.e.}, a single convolutional layer alone consumes a half LUT resource.

\begin{table}[htbp]
\centering
\caption{The resource report of Kintex-7 XC7K325T FPGA}
\begin{tabular}{@{}l|ccc@{}}
\hline
Resources & Used resources & Available resources & Resource occupancy\\ \hline
LUT       & 151637         & 203800              & 74.40\%                  \\
LUTRAM    & 936            & 64000               & 1.46\%                   \\
FF        & 180099         & 407600              & 44.19\%                  \\
BRAM      & 259            & 445                 & 58.20\%                  \\
DSP       & 185            & 840                 & 22.02\%                  \\
IO        & 6              & 500                 & 1.20\%                   \\
BUFG      & 3              & 32                  & 9.38\%                   \\
PLL       & 1              & 10                  & 10.00\%                  \\ \hline
\end{tabular}
\label{tab:resource}
\end{table}

Moreover, we record the model inference time and power consumption on both the CPU and the FPGA, where the Python project is deployed on a portable Intel NUC Mini PC. As presented in Tab. \ref{tab:power}, the FPGA demonstrates an inference time that is approximately $1/201$ of the CPU's and a power consumption that is nearly $1/42$ of the CPU. This significant reduction is attributed to the flexibility of FPGAs in processing CNN operations to enhance efficiency. Additionally, FPGA employs parallel computing techniques, with the advantages of low power consumption and accelerated computing rates. Such low power consumption also makes online bearing fault monitoring and diagnosis possible.

\begin{table}[htbp]
\centering
\caption{The inference time and power consumption of BearingPGA-Net on CPU and FPGA for 1000 samples.}
\begin{tabular}{@{}|l|c|c|@{}}\hline
Hardware & Inference time & Power Consumption  \\ \hline
Intel i5-1135G7@2.40GHz      & 1160us         & 28W               \\
\hline
Kintex-7 XC7K325T@100MHz     & 5.77us         & 0.67W     \\ \hline       
\end{tabular}
\label{tab:power}
\vspace{-0.5cm}
\end{table}

\section{Conclusions}

In this paper, we have proposed a lightweight and deployable CNN, referred to as BearingPGA-Net. By integrating FFT and DKD, BearingPGA-Net outperforms other state-of-the-art lightweight models when signals are noisy. Additionally, we have fully utilized the computational resources of the Kintex-7 FPGA and designed CNN accelerating scheme. The parallelism and model reuse significantly improves the running speed of our method, while the customized parameter quantization greatly alleviates the performance drop. Our deployment scheme has achieved over 200 times faster diagnosis speed compared to CPU and maintained over 97\% classification F1 score on the HIT bearing dataset. Finally, our model introduces a new approach to realizing online bearing fault diagnosis.

\bibliographystyle{IEEEtran}
\bibliography{bio.bib}

\vfill

\end{document}